\documentclass[pdflatex,sn-mathphys-num]{sn-jnl}

\usepackage{graphicx}%
\usepackage{multirow}%
\usepackage{amsmath,amssymb,amsfonts}%
\usepackage{amsthm}%
\usepackage{mathrsfs}%
\usepackage[title]{appendix}%
\usepackage{xcolor}%
\usepackage{textcomp}%
\usepackage{manyfoot}%
\usepackage{booktabs}%
\usepackage{algorithm}%
\usepackage{algorithmicx}%
\usepackage{algpseudocode}%
\usepackage{listings}%
\usepackage{comment}
\usepackage{color, soul}
\usepackage{url}

\newcommand*\rot{\rotatebox{90}}

\newcommand{\bk}[1]{}

\algnewcommand{\Inputs}[1]{%
  \State \textbf{Inputs:}
  \Statex \hspace*{\algorithmicindent}\parbox[t]{.8\linewidth}{\raggedright #1}
}
\algnewcommand{\Initialize}[1]{%
  \State \textbf{Initialize:}
  \Statex \hspace*{\algorithmicindent}\parbox[t]{.8\linewidth}{\raggedright #1}
}

\theoremstyle{thmstyleone}%

\theoremstyle{thmstyletwo}%

\theoremstyle{thmstylethree}%

\begin{document}

\title{Incremental Graph Construction Enables Robust Spectral Clustering of Texts}

\author[1]{\fnm{Marko} \sur{Pranjić}}
\equalcont{These authors contributed equally to this work.}

\author*[1]{\fnm{Boshko} \sur{Koloski}}\email{boshko.koloski@ijs.si}
\equalcont{These authors contributed equally to this work.}

\author[1,2]{\fnm{Nada} \sur{Lavra\v{c}}}
\author[1]{\fnm{Senja} \sur{Pollak}}
\author[2]{\fnm{Marko} \sur{Robnik-\v{S}ikonja}}

\affil[1]{\orgname{Jo\v{z}ef Stefan Institute and International Postgraduate School}, \city{Ljubljana}, \country{Slovenia}}
\affil[2]{\orgname{University of Ljubljana, Faculty of Computer and Information Science}, \city{Ljubljana}, \country{Slovenia}}

\abstract{Neighborhood graphs are a critical but often fragile step in spectral clustering of text embeddings. On realistic text datasets, standard $k$-NN graphs can contain many disconnected components at practical sparsity levels (small $k$), making spectral clustering degenerate and sensitive to hyperparameters. We introduce an incremental $k$-NN graph construction algorithm that preserves connectivity by design: each new node is linked to its $k$ nearest previously inserted nodes, which guarantees a connected graph for any $k$. We provide an inductive proof of connectedness and discuss implications for incremental updates when new documents arrive. We validate the approach on spectral clustering of SentenceTransformer embeddings using Laplacian eigenmaps across six clustering datasets from the Massive Text Embedding Benchmark. Compared to standard $k$-NN graphs, our method outperforms in the low-$k$ regime where disconnected components are prevalent, and matches standard $k$-NN at larger $k$.}
\maketitle

\section{Introduction}
\label{sec:introduction}

Graph-based machine learning has achieved strong results in domains with inherent relational structure, including protein interactions\cite{jha2022prediction}, materials science\cite{reiser2022graph}, and travel-time estimation\cite{derrow2021eta}.
Recently, graph methods have been extended to data without explicit graph structure, such as tabular data\cite{margeloiu2024gcondnet}, text\cite{iclr2023bk}, and medical imaging\cite{zaripova2023graph}, by constructing a neighborhood graph in which data points become nodes and edges encode pairwise similarity.

The two most common construction methods are $\epsilon$-threshold graphs and $k$-nearest neighbor graphs ($k$-NN). Both rely on a distance metric: the former connects all pairs within distance $\epsilon$, and the latter connects every node to its $k$ closest neighbors.
While sparser graphs are generally preferred for computational and memory efficiency, reducing $k$ or $\epsilon$ increases the risk of producing \emph{disconnected} graphs\cite{laplacian-eigenmaps}.
This risk comes from the fact that both methods use only \emph{local} proximity information and neither guaranty \emph{global} connectivity.
Theoretical analysis confirms this concern: a $k$-NN is asymptotically connected with high probability only when $k \geq 5.1774 \cdot \log N$\cite{10.1023/B:WINE.0000013081.09837.c0}, which already exceeds $k=30$ for as few as $N=300$ data points---well above the values typically used in practice.

Disconnected components are problematic for a range of downstream tasks. For example, in spectral clustering, each connected component can only be assigned to a single cluster; when the number of components equals or exceeds the number of desired clusters, the clustering becomes trivial, and no similarity-based criterion can improve it. In addition to clustering, disconnected graphs negatively affect other problems, such as item-based recommender systems, because unreachable items cannot be recommended \cite{10.1145/1639714.1639778}.

\begin{figure}[!hbt]
    \centering
    \includegraphics[width=0.95\textwidth]{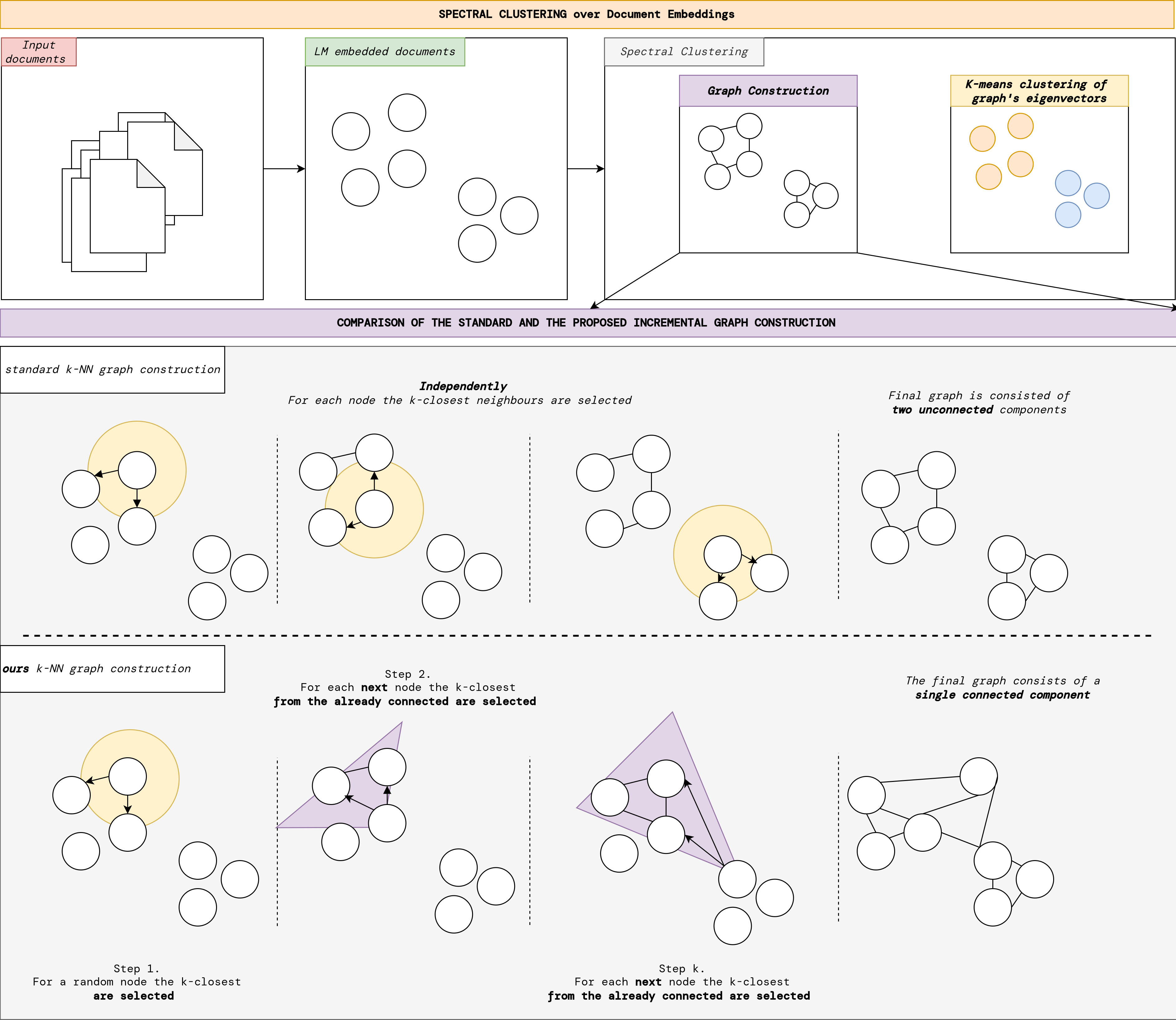}
   \caption{Proposed methodology. \textbf{Top}: General spectral clustering pipeline. After embedding the documents, a graph is constructed, projected into eigenspace, and clustered using k-means. \textbf{Bottom}: Comparison of a standard nearest-neighbor graph (which may be disconnected) with the proposed incremental approach that progressively links nodes to form a fully connected graph. The 
   colored triangle denotes the search for nearest neighbors among previously considered points, while the yellow circle represents the search for global nearest neighbors.}
    \label{fig:proposed-merhod}
\end{figure}

In this work, we address this fundamental limitation by proposing a simple modification to the standard $k$-NN based graph construction that \textbf{guarantees a connected graph for any value of~$k$}, for spectral clustering of texts. The method proceeds incrementally: nodes are added one at a time, and each new node is connected to its $k$-nearest neighbors among the nodes already in the graph, ensuring that every insertion preserves connectivity.
We introduce the graph-construction setting and illustrate the disconnection problem on the 20\,Newsgroups dataset.
We evaluate the resulting graphs on a downstream spectral clustering task using Laplacian eigenmaps\cite{laplacian-eigenmaps} and compare against the standard $k$-NN across two variants of six text-classification datasets. Our approach is summarized in Figure~\ref{fig:proposed-merhod}.

The paper is organized as follows.
Section~\ref{sec:related-work} reviews related work.
Section~\ref{sec:show-disconnected-knn} provides a motivating example and quantifies disconnected components on realistic text data.
Section~\ref{sec:inc-neig-graphs} presents the proposed incremental graph construction algorithm.
Section~\ref{sec:exp-setting}  describes the experimental setup and Section~\ref{sec:results} presents the experimental results. Section~\ref{sec:conclusion} concludes by presenting the directions for future work.

\section{Background and Related work}
\label{sec:related-work}
\label{sec:graphConstruction}

There are several ways to construct a graph representing the data in some metric space, where close nodes shall be connected, encoding the metric structure in graph connections. Two most popular ways are neighborhood graphs constructed from $\epsilon$-distance and from $k$-nearest neighbors, as described in \cite{laplacian-eigenmaps}. Constructing a graph from $\epsilon$-neighborhood requires a choice of distance parameter $\epsilon$: any two nodes with a distance less than $\epsilon$ are connected (also called $\epsilon$-radius). The downside is that connectedness is sensitive to $\epsilon$: too small yields disconnected components, while too large yields many connections that can impact running time and downstream performance. For $k$-NN, every node is connected to its $k$ nearest neighbors; this requires choosing $k$ and tends to produce connected graphs, but loses the distance-based geometric interpretation.

\textbf{Challenges with high-dimensional embeddings.} Deep-learning derived embeddings often use several hundred to several thousand dimensions, which introduces challenges to their analysis. In order to compare objects represented with dense embeddings, the usual choice is cosine distance, while Euclidean distance is avoided: cosine distance is not a proper metric (triangle inequality does not hold), and Euclidean distance in high-dimensional spaces loses geometric interpretation and exhibits unintuitive phenomena\cite{10.1007/3-540-44503-X_27}. Cosine distance scales distances to a range with maximum 2, making the $\epsilon$ parameter more sensitive to small changes. Theoretical results \cite{10.1007/3-540-49257-7_15} show poor discrimination between nearest and furthest points in high dimensions\cite{10.1007/3-540-44503-X_27}, making $\epsilon$-based selection unstable; $k$-NN was meant to address the difficulty of choosing $\epsilon$\cite{laplacian-eigenmaps}. However, both strategies may still generate multiple disconnected components, which is highly problematic for graph-based clustering.

\textbf{Neighborhood graphs and applications to clustering.} Representing relationships as graph connections transforms complex geometry to graph topology\cite{doi:10.1126/science.290.5500.2319} and enables tools from graph theory and complex network analysis \cite{doi:10.1007/s41109-019-0248-7}. Neighborhood graphs based on $\epsilon$-distance and $k$-NN connect points by a distance threshold or to their $k$ nearest neighbors. A more recent method is continuous $k$-NN (Ck-NN)\cite{doi:10.3934/fods.2019001}, but it can only be induced on a proper distance metric; in contrast, our work tries to induce a neighborhood graph even when a proper metric is not usable.

Among global connectivity-based approaches, popular methods include MST-based constructions such as $k$NN-MST\cite{knn-mst} and variants like PMST\cite{NIPS2004_dcda54e2} and RMST\cite{6737046}. PMST relies on Euclidean distances and random shifts, making it applicable only to low-dimensional data, while RMST performs comparisons and arithmetic on weights that are unintuitive under cosine distance. Although $k$NN-MST is a simple extension of $k$-NN by adding an MST, it efficiently resolves disconnected components and does not assume metric space or low-dimensional features. In contrast to the above, our method is applicable to undirected graphs, works in both low- and high-dimensional spaces, and is conceptually simple: it uses only $k$NN operations (no global features), enabling simpler implementation and faster processing.

\section{Demonstrating disconnected components on realistic data}
\label{sec:show-disconnected-knn}
We demonstrate the problems with neighborhood graphs constructed from textual embeddings on the well-known 20Newsgroups dataset\footnote{Available from \url{http://qwone.com/~jason/20Newsgroups/}}
collection of approximately 20,000 documents grouped into 20 different topics. We use this data as a motivating example to construct neighborhood graphs, studying different parameters and the numbers of disconnected components. The examples in this section use 7,532 documents that are part of the test set. We apply the recommended preprocessing: removal of headers, signature blocks, and quotation blocks. The preprocessed documents are encoded using the SentenceTransformers \cite{reimers-gurevych-2019-sentence} model (\texttt{all-MiniLM-L12-v2}) that generates 384-dimensional embeddings.
We use the cosine distance on these vectors to induce both $\epsilon$-distance and $k$-NN neighborhood graphs.

\subsection{Cosine distance $\epsilon$-neighborhood graph}
In order to analyze the $\epsilon$-neighborhood strategy, we find the minimal $\epsilon$ distance ($\epsilon_0$) that ensures connectedness and test several properties of the resulting graph. In order to determine  $epsilon_0$, we calculate the cosine similarity between all pairs from the datasets and apply a bisection algorithm over the range of parameter $epsilon$. This determined a constant $\epsilon_0=0.7694$, specific for this dataset and the model used to encode the data. This distance corresponds to a Cosine similarity of $(1-\epsilon_0) = 0.2306$. Table \ref{tab:20ng-conn-components-eps} presents some properties of this graph, the information on the $\epsilon$ distance used to infer the graph, the number of components in the resulting graph, the size of the largest component, the number of edges in this graph and the number of edges in a symmetric (undirected) graph. In addition to properties of a graph induced with $\epsilon_0$ distance, we provide information on graphs using $\epsilon$ that differ by 5\% and 10\% from $\epsilon_0$.


\begin{table}[t!]
\label{tab:20ng-conn-components-eps}
\centering
\begin{tabular}{cccrr}
\hline
{ $\epsilon$ } & Num comp. & Max comp. size & Graph edges & Digraph edges \\
\hline
$\epsilon_0 - 10\%$ & 19 & 7511 & 404,956 & 809,912 \\
$\epsilon_0 - 5\%$ &  3 & 7530 & 636,040 & 1,272,080 \\
$\epsilon_0$ & 1 & 7532 & 1,009,318 & 2,018,636 \\
$\epsilon_0 + 5\%$ & 1 &  7532 & 1,622,168 & 3,244,336 \\
$\epsilon_0 + 10\%$ & 1 & 7532 & 2,645,373 & 5,290,746 \\
\hline
\end{tabular}
\caption{Properties of $\epsilon$ neighborhood graphs for different values of $\epsilon$. A minimal number of graph edges to satisfy the connectedness of the graph is achieved with $\epsilon=\epsilon_0$. A graph contains disconnected components if the number of components is greater than 1.}

\end{table}

\begin{table}[b!]
\label{tab:20ng-conn-components}
\centering
\begin{tabular}{cccrr}
\hline

{ $k$ } & Num comp. & Max comp. size & Graph edges & Digraph edges \\
\hline
1 & 1,505 & 216 & 7,532 & 12,054 \\
2 &  49 & 7,342 & 15,064 & 23,470 \\
3 & 6 & 7,507 & 22,596 & 34,734 \\
4 & 2 &  7,527 & 30,128 & 45,978 \\
5 & 1 & 7,532 & 37,660 & 57,132 \\
\hline
\end{tabular}
\caption{Properties of $k$-NN neighborhood graphs for different values of parameter $k$. A graph contains disconnected components if the number of components is greater than 1.}

\end{table}

A graph generated from this dataset with a minimal distance required to retain graph connectivity requires around 1 million graph edges. For the calculation of Laplacian eigenmaps, we are interested in a graph with a sparse symmetric adjacency matrix. Such a matrix would contain 2 million entries (non-zero elements). The sparsity of the adjacency matrix is desired as it enables us to encode the relations within data using less information and speed up any subsequent processing step. From Table \ref{tab:20ng-conn-components-eps}, it can easily be seen that using $\epsilon$ only 5\% larger than the minimal required increases the number of edges by more than 60\%.

\subsection{Cosine-distance based $k$-nearest neighbor graph}
\label{sec:kNNraph}
Using the same setting as for the $\epsilon$-neighborhood graph, we generate $k$-NN neighborhood graphs by increasing the parameter $k$ and observing the number of connected components. A summary of results is presented in Table \ref{tab:20ng-conn-components}. For each row, the table includes the value of parameter $k$, the number of components, the number of nodes in the largest connected component, the number of connected pairs, and the number of edges in a symmetric adjacency matrix.
The fully connected directed graph requires $k=5$ nearest neighbors and contains more than 57,000 edges.
The $k$-nearest neighbors strategy was easier to tune to get the connected graph. The resulting graph is much sparser compared to the $\epsilon$-neighborhood strategy, containing less than 3\% of its number of edges. For many downstream tasks, this sparseness would translate into significantly faster processing.

\section{Incremental $k$-NN neighborhood graph construction}
\label{sec:inc-neig-graphs}

In this section, we describe the proposed modified $k$-NN neighborhood graph construction algorithm with two important features. First, it avoids disconnected components, a problem described in detail in Section \ref{sec:show-disconnected-knn}. Second, it allows the addition of new nodes after the graph is built. This opens up new possibilities for incrementally constructed graphs.

The pseudocode of the proposed incremental $k$-NN neighborhood graph construction algorithm is outlined in Algorithm \ref{alg:iter-knn}. The algorithm receives as the input the same parameters as a regular $k$-NN neighborhood graph construction algorithm, i.e. a set of $N$ vectors and parameter $k$. It returns a generated graph, analogous to the $k$-NN neighborhood graph, described in Section \ref{sec:kNNraph}.
The algorithm sequentially processes each input vector. For each input, it considers only nodes already present in the graph for $k$-nearest neighbors. Nodes in the graph are denoted with the set of vertices $V$ and connections between them $E$.

\begin{algorithm}
  \caption{Incremental $k$ Nearest Neighborhood Graph Construction}
  \label{alg:iter-knn}
  \begin{algorithmic}[1]
    \Inputs{$X=\{x_i;  ~ i\in 1,\ldots,N$\} \Comment{N input vectors} \\
    $k \in \{1 \ldots N-1\}$ \Comment{the number of nearest neighbors} }
    \Initialize{\strut$V \gets \{ x_1 \ldots x_k \}$ \Comment{ Initial set of $k$ vertices } \\
                      $E \gets \emptyset$ \Comment{ Initial empty set of edges }}
    \For{t = k+1 to N}  \label{lineno:for} \Comment{ Iteratively add the remaining vertices. }
      \State $Q \gets k$-NN$(x_t, V, k)$ \label{lineno:knn} \Comment{ Find $k$ nearest neighbors of $x_t$ in $V$. }
      \State $V \gets V \cup \{x_t\}$  \Comment{ Add $x_t$ to $V$. }
      \For{$q \in Q$}  \Comment{ Add edges from $x_t$ to all the nearest nodes in $Q$. }
        \State $E \gets E \cup \{(q, x_t)\}$
      \EndFor
    \EndFor\\
    \Return $G(V, E)$ \Comment{ Graph with $k (N-k)$ edges. }
  \end{algorithmic}
\end{algorithm}

A difference between constructing a regular $k$-NN neighborhood graph and an incremental Algorithm \ref{alg:iter-knn} can be seen in Line \ref{lineno:knn} of the algorithm.
While the standard $k$-NN neighborhood graph is constructed by searching for $k$ nearest neighbors among all possible nodes, Algorithm \ref{alg:iter-knn} considers only nodes already added to the graph.

\textbf{Properties of the algorithm.} When compared to the regular $k$-NN neighborhood graph construction based on an exact $k$-NN search, the construction of a graph with the proposed algorithm requires fewer comparisons. This is due to Algorithm \ref{alg:iter-knn} performing search only on a restricted set of nodes instead of the full set of nodes as with standard $k$-NN algorithm.

\textbf{Theorem.} The graph produced by the incremental step is necessarily connected. 
\begin{proof}
     The proof is by induction on the number of iterations. \\
    \textit{The base case.} The initial set of $k$ vertices has no connections. Adding vertex $x_{k+1}$, connects it with its $k$ nearest neighbors, i.e. with all $k$ existing nodes, and creates a single connected component. \\
    \textit{Inductive step.}  Adding new vertex $X_i, ~i>k+1$ to the connected graph, connects it with $k$ vertices in the connected component, and therefore extends the connected component by the newly added vertex. \\
    \textit{Consequence.} The final graph of $N$ vertices contains a single connected component.
\end{proof}

In the standard $k$-NN neighborhood graph, the consequence of adding a single additional node to the already constructed graph may be a large reconfiguration, as many existing nodes can be connected with the newly added node. This prevents efficient incremental expansion with new nodes and requires graph recreation.

In contrast, graphs constructed with Algorithm \ref{alg:iter-knn} can be expanded efficiently. This is due to the fact that the newly added node induces only local changes in the graph adjacency matrix. Those changes are restricted to the row and column corresponding to the newly added node. This can be exploited in applications where all nodes are not available at the start, like in the processing of streaming data.

\textbf{The weights of the adjacency matrix.} Laplacian Eigenmaps algorithm interprets entries in the graph adjacency matrix as the affinity between nodes. The affinity on the scale $[0,1]$ describes a spectrum of node closeness where 0 and 1 represent very distant and very close nodes. In \cite{laplacian-eigenmaps}, two schemes  for the choice of affinities are proposed, one based on the presence of connections between nodes and the other taking into account the distances between nodes.

\textbf{Connection-based node affinity.} A simple choice to construct the affinity matrix is to assign 1 to connections between nodes within $k$-NN neighborhood and 0 otherwise. The affinity matrix for the Laplacian eigenmaps has to be symmetric, while $k$-NN neighborhood graphs may have asymmetrical adjacency matrices. To remedy this, a popular approach is to average the adjacency matrix with its own transpose.

\textbf{Gaussian kernel-based node affinity.} Elements of affinity matrix can also be computed based on the kernel function defined as
$ \alpha_{ij} = e^{ - \frac{{|| x_i - x_j ||}^2} {4t}} $
with an additional hyperparameter $t \geq 0$. This variant of affinity matrix is inherently symmetric and is suitable for graphs based on the $\epsilon$-distance while the connectivity-based affinity matrix is usually used with $k$-NN neighborhoods.

\section{Experimental setting}
\label{sec:exp-setting}
Manifold learning approaches, including Laplacian eigenmaps, lack a natural measure to assess the quality of learned embedding. Existing research provides an answer for the evaluation of learned manifold only in the setting without noise \cite{ZHANG2012251}. However, the Laplacian eigenmaps have a natural application to  clustering that can be used to verify that embeddings are encoding relevant information. In our evaluation, we address the following aspects. First, we verify that the proposed approach doesn't introduce large performance regression when compared to the standard $k$-NN neighborhood graph. Second, we test if our approach improves the clustering results in a restricted setting, using small $k$, where disconnected components would otherwise occur. Third, we gain insight into the amount of information lost when compared to the original high-dimensional embeddings. Finally, due to the inherent dependence on the ordering of nodes in Algorithm \ref{alg:iter-knn} on a resulting graph, we are interested in the stability of the approach with respect to the ordering of the nodes.

We address the first two goals by running clustering, configured to use our approach and using a $k$-NN neighborhood graph.
With a low value of parameter $k$, we expect more disconnected components and a larger difference to our approach. With higher values of $k$, both clustering evaluations will likely run on a connected graph. In this scenario, our approach does not benefit the task and may negatively influence the results due to fewer available nodes. This will enable us to quantify the negative aspects of our approach when used on a dataset that already produces connected neighborhood graphs.
Although the amount of information lost by encoding high-dimensional embeddings to low-dimensional ones through Laplacian eigenmaps cannot be measured directly, we can evaluate the full-graph and our incremental approach on the same downstream task and quantify the difference in performance. In particular, we compare the clustering performance of low-dimensional embeddings produced by our approach with the clustering performance of original high-dimensional embeddings using the K-means algorithm.

To measure variance, due to the input node order, we run spectral clustering using a neighborhood graph generated by Algorithm \ref{alg:iter-knn} for each dataset ten times, each time randomizing the ordering of nodes, using connection-based node affinity. Due to the guaranteed connectedness of the graph, the first dimension of all embedding vectors is collapsed to a constant value, so we use an additional Laplacian eigenvector instead. The QR factorization algorithm used for clustering of spectral embeddings was likewise amended to ignore the first (constant-value) dimension when assigning cluster labels to the embedding vectors.

In the following subsections, we provide a short overview of the evaluation datasets, methods used to infer document representations, neighborhood graph construction, and clustering metrics. The results of experiments are discussed in Section \ref{sec:results}.

\subsection{Datasets}
We evaluate our approach on clustering datasets included in the Massive Text Embedding Benchmark \cite{mteb} (MTEB)\footnote{Although MTEB comes with its own evaluation based on MiniBatchKMeans clustering, we do not use it for our purposes and evaluate our approach using spectral clustering with QR decomposition.}.
The datasets come from six sources,
and are assembled in two variants; the first variant contains only titles, and the second variant of a dataset uses a concatenation of the title and body of a document. In MTEB, these two variants are called sentence-to-sentence (S2S) and paragraph-to-paragraph (P2P), respectively.
Each dataset is partitioned into several sets, where each set contains a subset of labels such that both fine- and coarse-grained differences are evaluated. In total, there are 182 different clustering problems. Detailed statistics on the datasets can be found in Table \ref{tab:data_table}. Additional details on the datasets are available in \cite{mteb}.

\begin{table}[t!]
    \centering
    \begin{tabular}{l|c c|c c c c | c c c c}
    \hline
      \multicolumn{3}{c}{}  & \multicolumn{4}{|c}{S2S} & \multicolumn{4}{|c}{P2P}   \\
      \hline
     Dataset & Title & Body & docs. & avg. len. & P &  C & docs. & avg. len. & P & C   \\
     \hline
      ArXiv& $\checkmark$& $\checkmark$ & 732723 &  74.00 & 31  & 171  & 732723 &  1009.90 & 31 & 171 \\
      BioRxiv& $\checkmark$ & $\checkmark$& 75000 & 101.60 & 10& 26  & 75000 & 1666.20  & 10   & 26 \\
      MedRxiv&  $\checkmark$ & $\checkmark$ & 37500 & 114.70 & 10 & 51 & 37500 & 1981.20  & 10 & 51 \\
      Reddit&  $\checkmark$ &$\checkmark$ & 420464 & 64.70  & 25 & 50  & 459399 & 727.70 & 10 & 87 \\
      StackExchange &  $\checkmark$ & $\checkmark$& 373850  & 57.00 & 25  & 50 & 75000 & 1090.70 & 10 & 385\\
      TwentyNewsgroups &  $\checkmark$ & -- & 59545   & 32.00  & 10 & 20 & --- & --- & ---& ---  \\
      \hline
    \end{tabular}
    \caption{An overview of text datasets used in the evaluation. We report the number of documents, average document length, number of partitions in column P, and maximum number of clusters within the problems in column C. The statistics are presented for both sentence-to-sentence (S2S) and paragraph-to-paragraph (P2P) tasks.}
    \label{tab:data_table}
\end{table}

\subsection{Document representation}
\label{sec:doc-repr}
SentenceTransformers \cite{reimers-gurevych-2019-sentence} is a unified framework for sentence, text and image embeddings that uses Siamese and triplet network models to produce semantically meaningful high-dimensional data representations. For all our experiments, we use text embeddings generated from \texttt{all-MiniLM-L12-v2} model. It encodes input text limited to 256 tokens to an embedding of 384 dimensions. Laplacian eigenmaps algorithm was used to transform those embeddings to a low-dimensional representation with dimensionality matching the number of clusters.

\subsection{Metrics}
As per MTEB \cite{mteb} benchmarking suite, the clustering performance is evaluated using \textit{V-measure} \cite{rosenberg-hirschberg-2007-v} ($V_1$) score. \textit{Homogeneity} $h$ is used to evaluate how close a given clustering is to an ideal clustering where each cluster only contains items of a single class.
\textit{Completeness} $c$ evaluates how close a given clustering is to perfectly complete clustering where all members of a single class are assigned to a single cluster (achieving completeness $c=1$). Putting all items into a single cluster would achieve $c=1$, but the \textit{homogeneity} would be zero ($h=0$).
A different type of degenerate clustering, assigning each item to a separate cluster, will have a perfect \textit{homogeneity} ($h=1$), but the \textit{completeness} will be zero. The harmonic mean of those two opposite aspects is called \textit{V-measure} ($V_1$), also known as Normalized Mutual Information (NMI). Similarly to the familiar F-score, it has a parameter $\beta$ where larger values give more weight to completeness.
\[ V_\beta = (1+\beta)\frac{h \cdot c}{\beta\cdot h + c} \]

For the baseline $k$-NN neighborhood graph, the evaluation also checks for any disconnected components in the neighborhood graph. This check is not needed for our incremental solution as the graph is guaranteed to be connected.

\section{Results}
\label{sec:results}
This section presents the results of spectral clustering and K-means clustering. For graph creation, we use our incremental approach or the standard $k$-NN neighborhood approachj.
Following \cite{mteb}, the results of different partitions from the same data source are averaged.
\subsection{Main Results}
\label{sec:main_res}
\begin{figure}[b!]
\includegraphics[width=\textwidth]{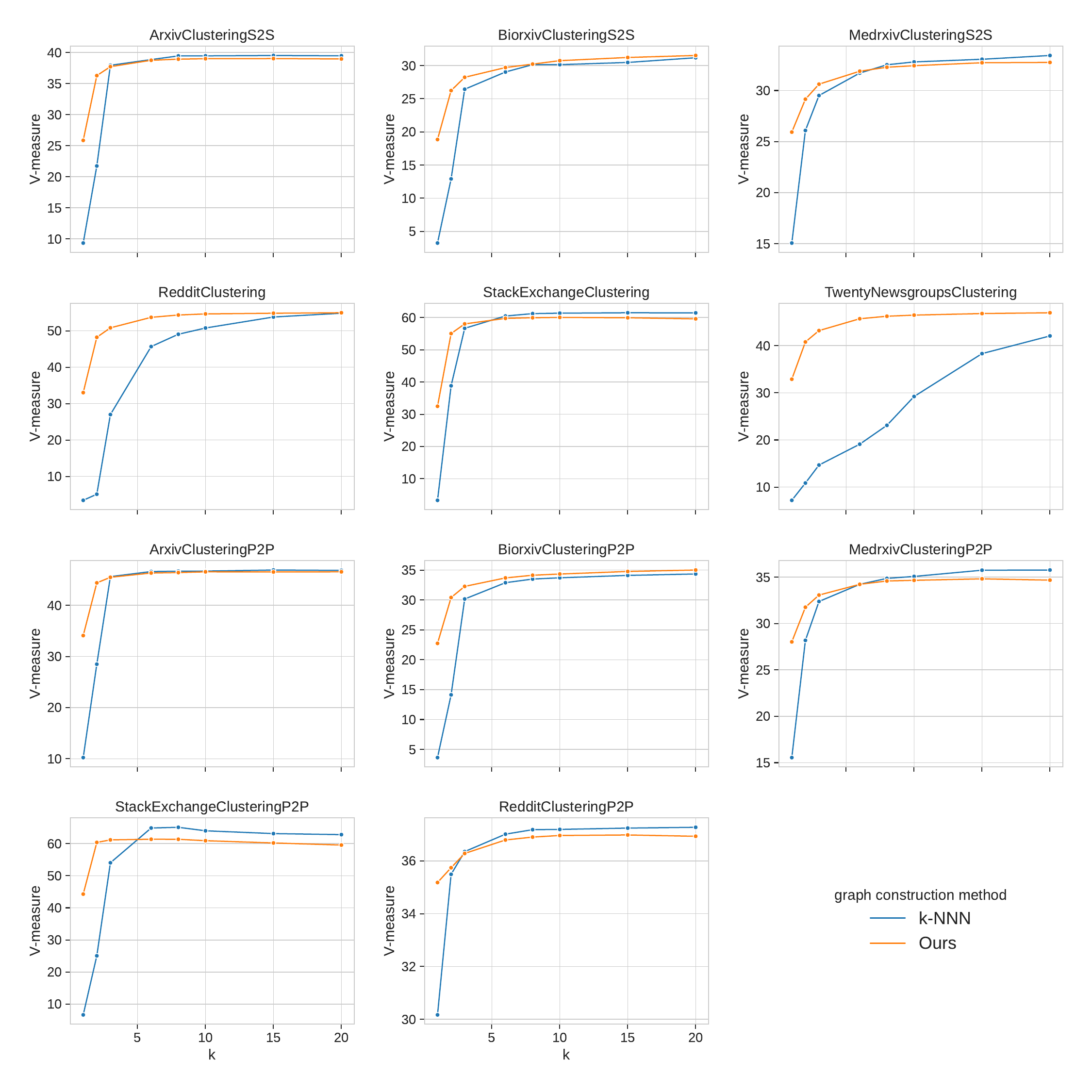}
\caption{Clustering performance of embeddings induced using incremental $k$-NN (Ours) and standard $k$-NN for a range of parameter $k$. Points, where the standard $k$-NN neighborhood induces connected graphs, are marked, while in Ours, the graph is always connected. Results for sentence-to-sentence (S2S) tasks are shown on the top and for paragraph-to-paragraph (P2P) tasks at the bottom.}
\label{fig:minilm-results-by-k}
\end{figure}

\begin{table}[ht]
\resizebox{\textwidth}{!}{
\begin{tabular}{cccccccc|ccccc}
\hline
{} & {} & \multicolumn{6}{c|}{S2S}  & \multicolumn{5}{|c}{P2P} \\
 \hline
{} & {k} &  Arx. &  Bio. &  Med. &  Red. &  SE &  20NG &  Arx. &  Bio. &  Med. &  Red. &  SE \\
\hline
\multirow{8}{*}{\rot{k-NN LD}} &
     1 & 9.33 & 3.24 & 15.07 & 3.43 & 3.31 & 7.19 & 10.23 & 3.63 & 15.55 & 6.67 & 30.17       \\
{} & 2 & 21.73 & 12.91 & 26.09 & 5.11 & 38.83 & 10.85 & 28.49 & 14.13 & 28.18 & 25.06 & 35.49 \\
{} & 3 & 37.93$\uparrow$ & 26.43 & 29.51 & 26.99 & 56.67 & 14.67 & 45.61$\uparrow$ & 30.16 & 32.37 & 54.03 & 36.35$\uparrow$ \\
{} & 6 & 38.85$\uparrow$ & 29.03 & 31.71 & 45.67 & 60.48$\uparrow$ & 19.10 & 46.60$\uparrow$ & 32.89 & 34.22-- & 64.86$\uparrow$ & 37.01$\uparrow$ \\
{} & 8 & 39.44$\uparrow$ & 30.14 & 32.52$\uparrow$ & 49.05 & 61.22$\uparrow$ & 23.10 & 46.67$\uparrow$ & 33.50 & 34.86$\uparrow$ & \bf{65.08}$\uparrow$ & 37.18$\uparrow$  \\
{} & 10 & 39.45$\uparrow$ & 30.14 & 32.80$\uparrow$ & 50.77 & 61.36$\uparrow$ & 29.22 & 46.69$\uparrow$ & 33.71 & 35.07$\uparrow$ & 63.99$\uparrow$ & 37.19$\uparrow$ \\
{} & 15 & \bf{39.51}$\uparrow$ & 30.47 & 33.06$\uparrow$ & 53.79 & \bf{61.49}$\uparrow$ & 38.32 & \bf{46.90}$\uparrow$ & 34.11 & 35.74$\uparrow$ & 63.12$\uparrow$ & 37.24$\uparrow$ \\
{} & 20 & 39.46$\uparrow$ & 31.18 & \bf{33.43}$\uparrow$ & 54.83 & 61.45$\uparrow$ & 42.07 & 46.83$\uparrow$ & 34.35 & \bf{35.76}$\uparrow$ & 62.80$\uparrow$ & \bf{37.27}$\uparrow$ \\
 \hline
\multirow{8}{*}{\rot{Ours LD}} &
     1 & 25.85$\uparrow$ & 18.85$\uparrow$ & 25.93$\uparrow$ & 33.02$\uparrow$ & 32.48$\uparrow$ & 32.89$\uparrow$ & 34.09$\uparrow$ & 22.74$\uparrow$ & 28.02$\uparrow$ & 44.28$\uparrow$ & 35.18$\uparrow$   \\
{} & 2 & 36.27$\uparrow$ & 26.23$\uparrow$ & 29.15$\uparrow$ & 48.22$\uparrow$ & 55.04$\uparrow$ & 40.79$\uparrow$ & 44.39$\uparrow$ & 30.41$\uparrow$ & 31.75$\uparrow$ & 60.37$\uparrow$ & 35.74$\uparrow$   \\
{} & 3 & 37.71 & 28.23$\uparrow$ & 30.62$\uparrow$ & 50.82$\uparrow$ & 58.03$\uparrow$ & 43.20$\uparrow$ & 45.48 & 32.27$\uparrow$ & 33.06$\uparrow$ & 61.16$\uparrow$ & 36.28   \\
{} & 6 & 38.74 & 29.70$\uparrow$ & 31.89$\uparrow$ & 53.70$\uparrow$ & 59.75 & 45.72$\uparrow$ & 46.32 & 33.69$\uparrow$ & 34.22-- & 61.36 & 36.79   \\
{} & 8 & 38.92 & 30.22$\uparrow$ & 32.28 & 54.34$\uparrow$ & 59.94 & 46.25$\uparrow$ & 46.40 & 34.14$\uparrow$ & 34.58 & 61.35 & 36.90   \\
{} & 10 & 38.99 & 30.74$\uparrow$ & 32.43 & 54.64$\uparrow$ & 60.05 & 46.49$\uparrow$ & 46.55 & 34.34$\uparrow$ & 34.65 & 60.91 & 36.96   \\
{} & 15 & 39.01 & 31.22$\uparrow$ & 32.72 & 54.83$\uparrow$ & 59.93 & 46.82$\uparrow$ & 46.52 & 34.77$\uparrow$ & 34.81 & 60.20 & 36.98   \\
{} & 20 & 38.95 & 31.52$\uparrow$ & 32.75 & \bf{54.96}$\uparrow$ & 59.60 & 46.99$\uparrow$ & 46.55 & 35.01$\uparrow$ & 34.67 & 59.55 & 36.93   \\
 \hline
\multicolumn{2}{c}{K-means HD} & 38.24 & \textbf{33.99} & 32.73 & 52.87 & 54.85 & \textbf{48.67} & 46.76 & \textbf{38.39} & 34.63 &  55.53 & 37.08\\
\hline
\end{tabular}}
\caption{$V_1$ scores (in percentages, larger is better) for spectral clustering of low-dimensional embeddings (LD) using standard $k$-NN neighborhood graph and our incremental approach. An arrow is shown to mark the relative difference between two variants for the same value of $k$. Additionally, we provide results for the clustering of high-dimensional embeddings (HD) using the K-means algorithm.}
\label{tab:results-means}
\end{table}

\begin{table}[ht]
\resizebox{\textwidth}{!}{
\begin{tabular}{cccccccc|ccccc}
\hline
{} & {} & \multicolumn{6}{c|}{S2S}  & \multicolumn{5}{|c}{P2P} \\
 \hline
{} & {k} &  Arx. &  Bio. &  Med. &  Red. &  SE &  20NG &  Arx. &  Bio. &  Med. &  Red. &  SE \\
\hline
\multirow{8}{*}{\rot{Ours LD}} &
1 & 0.84\% & 0.83\% & 0.52\% & 1.25\% & 1.42\% & 1.42\% & 1.06\% & 0.82\% & 0.55\% & 1.38\% & 0.21\%  \\
{} & 2 & 0.57\% & 0.77\% & 0.52\% & 1.03\% & 0.95\% & 1.33\% & 0.55\% & 0.66\% & 0.48\% & 0.86\% & 0.17\%  \\
{} & 3 & 0.45\% & 0.71\% & 0.45\% & 0.92\% & 0.81\% & 1.53\% & 0.47\% & 0.53\% & 0.40\% & 0.75\% & 0.15\%  \\
{} & 6 & 0.43\% & 0.59\% & 0.38\% & 0.63\% & 0.88\% & 0.86\% & 0.36\% & 0.55\% & 0.41\% & 0.60\% & 0.14\%  \\
{} & 8 & 0.34\% & 0.43\% & 0.41\% & 0.48\% & 0.86\% & 0.99\% & 0.34\% & 0.48\% & 0.38\% & 0.53\% & 0.14\%  \\
{} & 10 & 0.32\% & 0.46\% & 0.39\% & 0.55\% & 0.74\% & 0.75\% & 0.28\% & 0.42\% & 0.42\% & 0.54\% & 0.15\%  \\
{} & 15 & 0.26\% & 0.49\% & 0.35\% & 0.47\% & 0.70\% & 0.80\% & 0.28\% & 0.46\% & 0.40\% & 0.50\% & 0.15\%  \\
{} & 20 & 0.25\% & 0.39\% & 0.42\% & 0.36\% & 0.64\% & 0.61\% & 0.26\% & 0.37\% & 0.38\% & 0.60\% & 0.14\%  \\
\hline
\end{tabular}}
\caption{Standard deviation of clustering on neighborhood graphs created by our approach.}
\label{tab:results-stdev}
\end{table}

Figure \ref{fig:minilm-results-by-k} and Table \ref{tab:results-means} show V-measure results comparing our incremental approach (\textit{Ours LD}) and standard $k$-NN neighborhood (\textit{k-NN LD}).
We use high-dimensional embeddings generated from the SentenceTransformer model described in Section \ref{sec:doc-repr} and apply K-means clustering in order to have a reference point for an upper bound of the clustering performance on the low-dimensional spectral embeddings. This value is shown in Table \ref{tab:results-means} under the row \textit{K-means HD}.
Standard clustering shows low performance with low values of $k$ and, in general, recovers most of the final score above $k=8$ when a further increase in $k$ brings only diminishing returns.
Spectral clustering using our neighborhood graph converges much faster to the limit of its performance. Already at $k=3$ the performance is close to the top-score achieved for a dataset. Overall, our incremental approach achieved consistently better scores across all measured values of $k$ on four out of nine datasets, while in others, it outperformed the standard approach for small values of $k$.

When comparing the two approaches to clustering, the largest differences in performance are seen on the  \textit{TwentyNewsgroups\footnote{TwentyNewsgroups is a variant of 20Newsgroups datasets from MTEB where only subject headers are available.}} dataset where our approach achieves higher score. We believe that the reason for this is a large number of disconnected components when using the standard $k$-NN neighborhood graph.
On every dataset evaluated, there was at least one split of the data with disconnected components using $k=5$. Some dataset partitions within \textit{TwentyNewsgroups} dataset exhibited disconnected components using $k=15$, and \textit{Reddit} (sentence-to-sentence) even with $k=20$.

Table \ref{tab:results-means} also provides a comparison with K-means clustering. One shall note an important difference between spectral and K-Means clustering, namely the spectral clustering uses Laplacian eigenmaps-based dimensionality reduction, and the clustering results show performance on the low-dimensional space. Results for K-means are obtained by clustering directly in the high-dimensional space.
By reducing the dimensionality of the data, one loses some information present in the high-dimensional space, and somewhat inhibited clustering performance is expected. As can be seen from Table \ref{tab:results-means}, this is not always the case, and for some datasets, our approach exhibits improvement in the clustering metrics. This can be explained by the ability of the spectral clustering method to perform well even with irregularly shaped clusters, while K-means has much stricter assumptions on the cluster shape.

The proposed incremental algorithm uses a given ordering of the graph nodes, and the resulting neighborhood graph is dependent on this ordering.
In Table \ref{tab:results-stdev}, we analyze the stability of the clustering performance with regard to the ordering of processed graph nodes. The table shows an average standard deviation for a dataset calculated from ten runs of the clustering algorithm. The results show that even with very low values of $k$, like $k=3$, the standard deviation of clustering performance is rarely above 1\% and is often below 0.5\%. We also show that increasing the value of $k$ reduces the standard deviation of results for the clustering task.
We also considered experiments with HDBSCAN \cite{McInnes2017} clustering, but they are not included due to the very low quality of results. The limitation of the reference implementation is the lack of support for non-metric distances like cosine distance, so Euclidean distance on normed vectors was used. We believe the reason for the inadequate results of HDBSCAN is the inappropriate use of metric distance in combination with high-dimensional vector space.

\subsection{Ablation study}
In this paper, we have so far only used the latent embedding space of the \textit{all-miniLM-L12} model. Although this model is robust, it has some downsides, such as the length of the context, the dimensionality of the embedding, and the number of parameters, making it an unsuitable candidate for larger and more complex documents. In Section \ref{sec:related-work}, we have however introduced several works that improve clustering results by adding information from the MST graph. We are therefore interested if adding this information further improves the proposed approach.

\subsubsection{The text embedding model}
\label{sec:lat_model}
To evaluate the effects of different embedding models and the dimensionality of the embedding space, we use several model variants, described in Table \ref{tab:models.}. Our aim is to verify if the approach described in this paper is sensitive to the quality of high-dimensional embeddings or if Laplacian eigenmaps transformation to low-dimensional vectors already saturates the expressive power of this low-dimensional embedding space.

\begin{figure}[t]
    \centering
    \resizebox{0.95\textwidth}{!}{\includegraphics{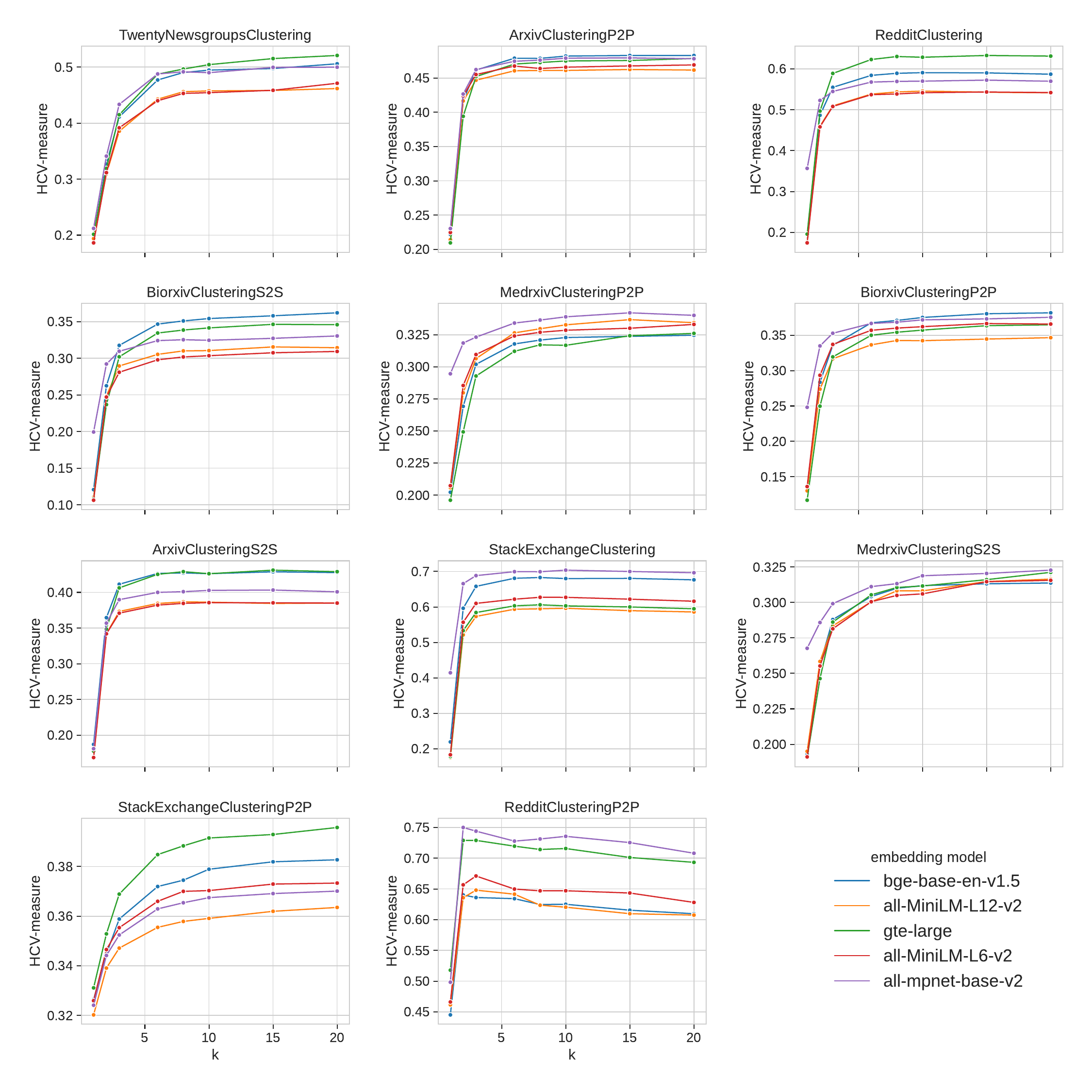}}
    \caption{The impact of different embedding models applied to different datasets.}
    \label{fig:multi_models}
\end{figure}

\begin{table}[t]
    \begin{tabular}{c|c|c|c|c}
    \hline
         Model & Emb. Dimensions. & Max tokens & Model Size (GB)   & average V-measure\\
         \hline
        all-MiniLM-L6-v2 \cite{reimers-gurevych-2019-sentence}& 	384	&256 &	0.09 & 0.41  \\
        all-MiniLM-L12-v2	\cite{reimers-gurevych-2019-sentence} & 384	& 256 &	0.13 &  0.40 \\
        all-mpnet-base-v2	\cite{reimers-gurevych-2019-sentence} & 768	& 384 &	0.42 & 0.42\\
        bge-base-en-v1.5 \cite{chen2024bge} & 768	&512 &	0.44 & \textbf{0.44} \\
        gte-large	\cite{li2023towards} & 1024 &	512	& 0.67 &\textbf{0.44} \\
        \hline
    \end{tabular}
    \caption{Additional models used to evaluate the properties of the text embedding space.}
    \label{tab:models.}
\end{table}

For each embedding model, we apply the proposed algorithm with the same setting as in the previous experiments and test it on all of the available datasets. In Figure~\ref{fig:multi_models} we present the results. Across datasets, we observed enhanced performance with increasing $k$-values, with the notable exception of the \textit{RedditClusteringP2P} dataset. This variance in model performance across datasets alludes to the complex relationship between dataset-specific features and the capability of models to leverage these features.

\begin{figure}[ht]
    \centering
    \resizebox{0.75\textwidth}{!}{\includegraphics{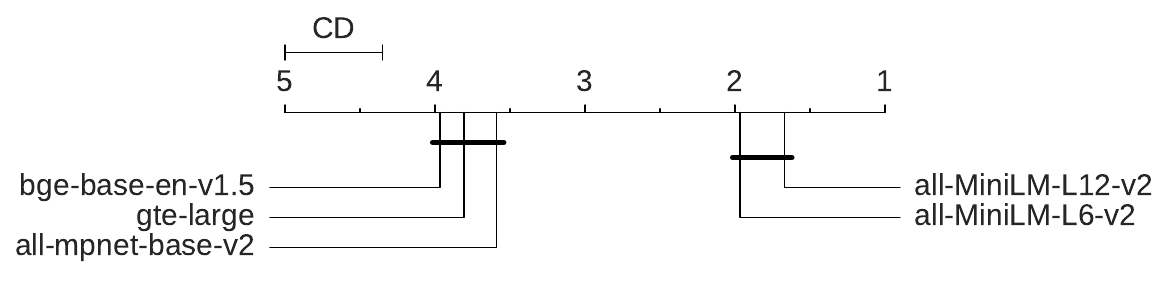}}
    \caption{The Nemenyi-Friedman test with critical distance of multiple embedding models across all datasets and values of $k$.}
    \label{fig:nemenyi}
\end{figure}

We next analyzed if there is a statistically significant difference between embedding models based on the Friedman rank test with Nemenyi \cite{demvsar2006statistical} post-hoc correction (refer to Figure \ref{fig:nemenyi}) at the standard $\alpha = 0.05$ significance level. We rank the models in increasing order of V-measure, therefore, higher rank means better performance. The test outcomes, yielding a p-value below 0.01 and a CD of 0.65, highlight significant performance disparities among larger models. Specifically, larger models (\textit{bge-base-en-v1.5}, \textit{gte-large}, \textit{all-mpnet-base-v2}) achieve higher V-measure scores compared to the smaller \textit{all-MiniLM-L12-v2} and \textit{all-MiniLM-L6-v2} models.

\subsubsection{Adding MST to the incrementally built graph}
\label{sec:mst}
In our incremental approach, we process nodes sequentially, building a connected graph using only local nearest neighbor information. In Section \ref{sec:main_res}, we compare our approach to a standard nearest neighborhood graph that also uses local nearest neighbor information in a greedy approach to graph construction with the downside of the graph being potentially unconnected. A minimum spanning tree (MST) and its variants are posed as a main alternative in the related work (Section \ref{sec:related-work}) and were shown to improve the clustering.

To determine if incorporating this global information affects the performance of our approach, we integrate the MST into graph constructed using Algorithm \ref{alg:iter-knn}. We compare the performance of incremental k-nearest neighbors (kNN) graphs (referred to as "\textit{base}") against the same kNN graph augmented with the MST (referred to as "\textit{base+mst}"), employing the Bayesian Signed Rank test following \citet{JMLR:v18:16-305}, with a Region of Practical Equivalence (ROPE) of 0.01. 

\begin{table}[ht]
    \centering
    \resizebox{0.80\textwidth}{!}{  \begin{tabular}{lrrr}
    \toprule
    Dataset & P(base $<$ \textit{base+mst}) & P(\textit{ROPE}) & P(\textit{base+mst} $>$ base) \\
    \midrule
    ArxivClusteringP2P & 0.427 & \textbf{0.573} & 0.000 \\
    ArxivClusteringS2S & \textbf{0.525} & 0.475 & 0.000 \\
    BiorxivClusteringP2P & 0.401 & \textbf{0.599} & 0.000 \\
    BiorxivClusteringS2S & 0.331 & \textbf{0.669} & 0.000 \\
    MedrxivClusteringP2P & 0.092 & \textbf{0.908} & 0.000 \\
    MedrxivClusteringS2S & 0.107 & \textbf{0.893} & 0.000 \\
    RedditClustering & 0.228 & \textbf{0.772} & 0.000 \\
    StackExchangeClustering & \textbf{0.685} & 0.315 & 0.000 \\
    StackExchangeClusteringP2P & 0.045 & \textbf{0.955} & 0.000 \\
    TwentyNewsgroupsClustering & \textbf{0.951} & 0.049 & 0.000 \\
    \bottomrule
    \end{tabular}}
    \caption{Bayesian assessment of model results comparing our approach with and without minimum spanning tree (MST) structure embedded within the graph using ROPE of 0.01.}
        \label{tab:bayes}
\end{table}

\begin{figure}[ht]
    \centering
    \resizebox{0.75\textwidth}{!}{\includegraphics{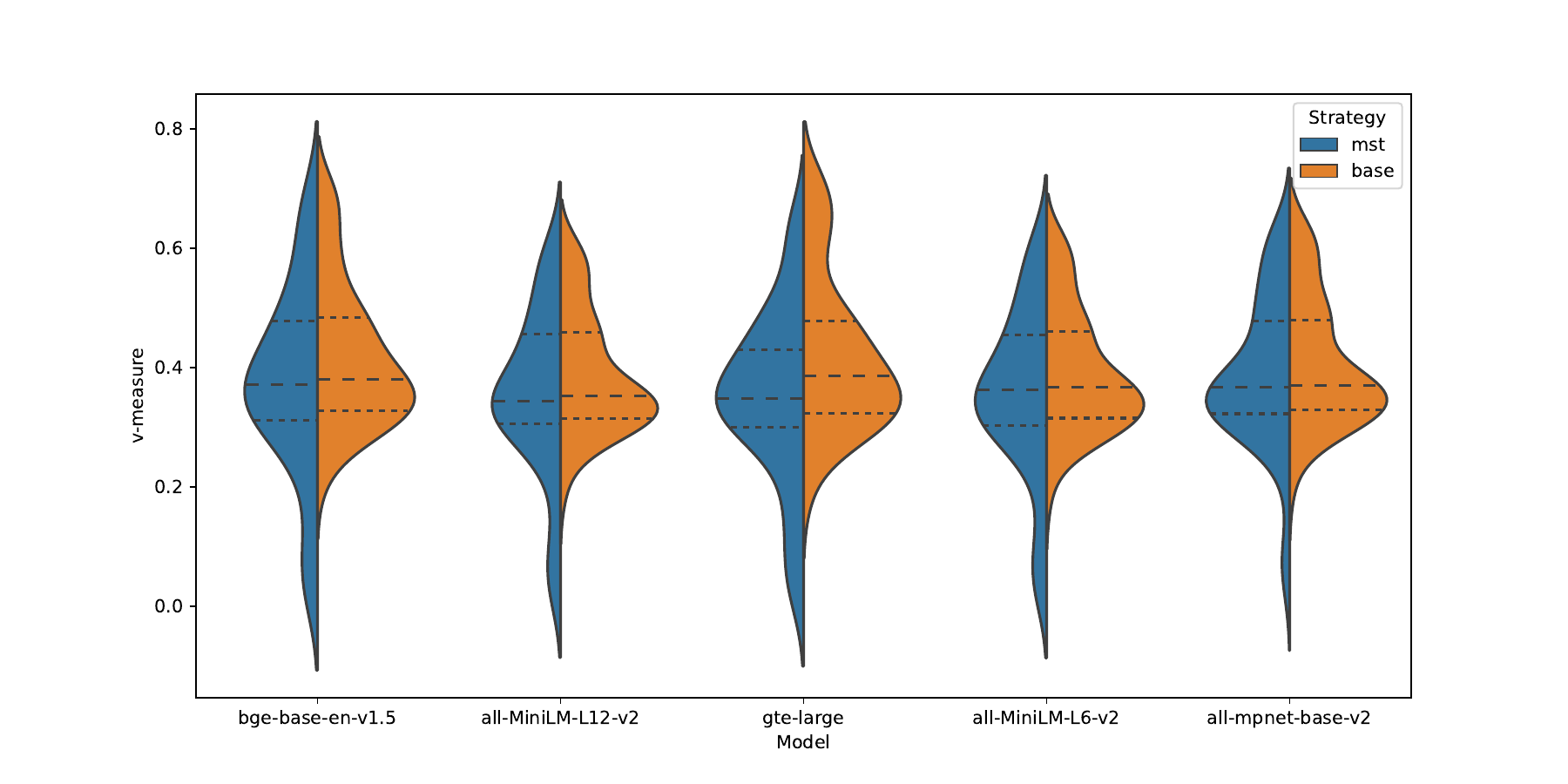}}
    \caption{Comparison of impact of strategy and model selection, averaged across datasets.}
    \label{fig:model_strategy-label}
\end{figure}

Results in Table~\ref{tab:bayes} tell that in some datasets, such as 'TwentyNewsgroupsClustering', the \textit{base} model outperforms the \textit{base+mst} model, suggesting that the addition of MST inhibits the clustering performance. Conversely, high probabilities in the Region of Practical Equivalence (ROPE) for multiple datasets, such as 'MedrxivClusteringP2P' and 'RedditClustering', indicate that the performance differences between the \textbf{base} and \textbf{base+mst} models are often not practically significant.
This result suggests that incremental $k$-nearest neighbor algorithm does not benefit from additional global features like those provided with MST, and including this information inhibits the performance on some datasets.
We note that the behavior described above is consistent across different latent space models (introduced in Section \ref{sec:lat_model}), with higher quality models showing a higher difference between means (see Figure \ref{fig:model_strategy-label}).

\begin{figure}[ht]
    \centering
    \resizebox{0.7\textwidth}{!}{\includegraphics{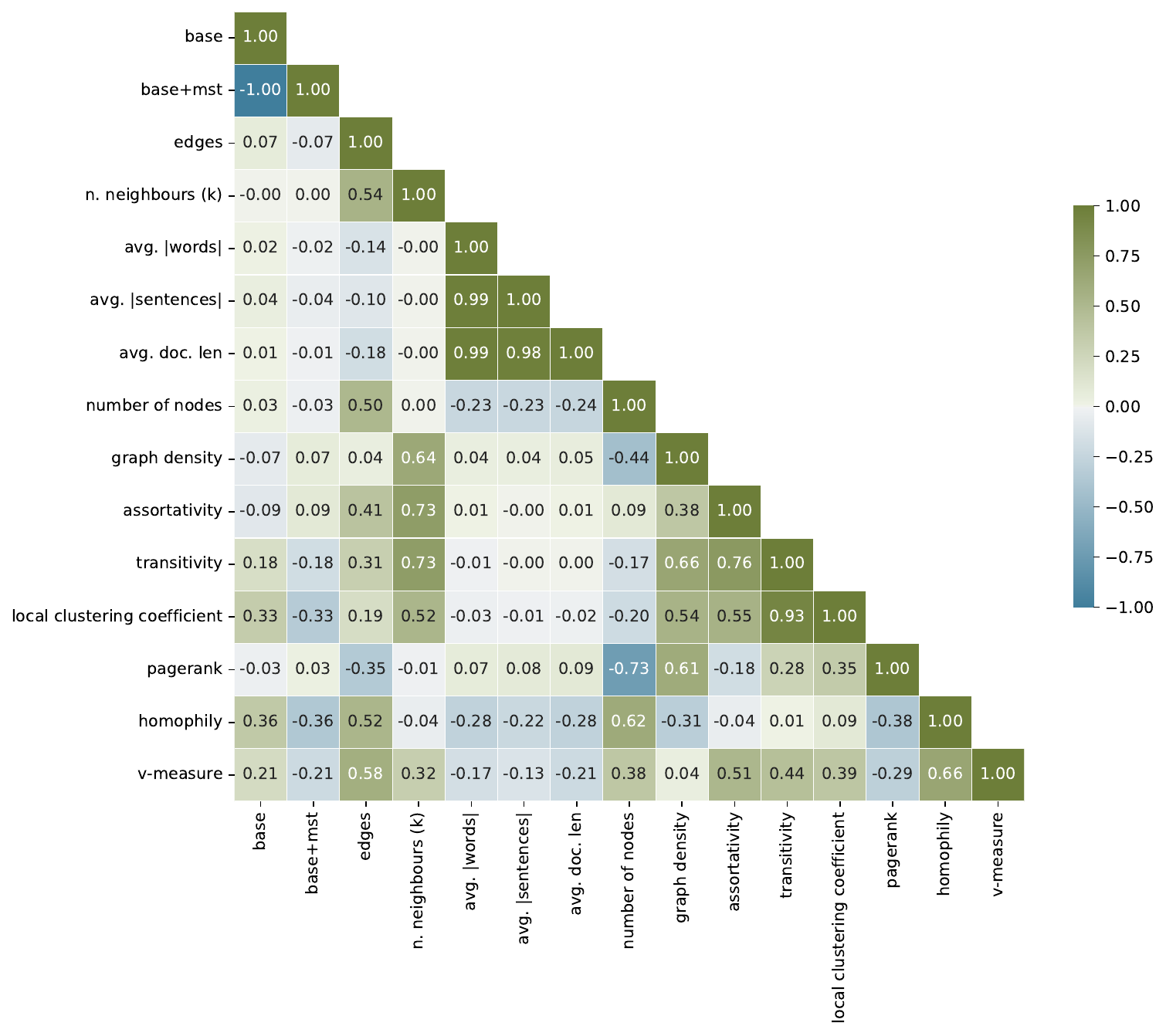}}
    \caption{Correlation of the graph-level and document-level properties and the \textit{V-measure}. }
    \label{fig:enter-label}
\end{figure}

\subsubsection{Comparing graph properties}
Next, we focus on differences between graphs produced by the proposed incremental algorithm (\textbf{base}) and those additionally enriched with an MST (\textbf{base+mst}).

We consider several graph- and node-level properties. Given a graph $G = (V, E)$, where $V$ is the set of documents and $E$ is the set of edges connecting vertices, we examine the following measures: graph density, assortativity, transitivity, local clustering coefficient, PageRank, and homophily. Definitions of these metrics are provided in Appendix~\ref{app:graph-statistics}. We also include document-level statistics: average number of words (avg.$ |$words$|$), average number of sentences (avg.\ $|$sentences$|$), and average number of characters (avg.\ doc.\ len). 

Computing exact graph statistics for large graphs is computationally intensive. Following related work \cite{graham2014empirical}, we therefore use a Monte Carlo simulation of graphs, selecting the number of nodes/edges examined such that statistical accuracy falls within a 95\% confidence interval to the second decimal place. For this analysis, we use \textit{all-mini-LM-v12} as the embedding model and compare both the original graph structure (\textit{base}) and the MST-enriched graph structure (\textit{base+mst}) (see Section~\ref{sec:mst}). We compute the correlation between these properties and the V-measure.

In Figure~\ref{fig:enter-label}, we can observe that the construction of the graph \textbf{base} weakly correlates (coefficient 0.21) with the V-measure. For the \textit{document-level statistic}, we find negative correlations of longer documents to V-measure, i.e. the number of characters (-0.21), words (-0.17), and sentences (-0.13). The larger graphs at both the \textbf{node} (0.38) and \textbf{edge} (0.58) level correlate positively with the V-measure, which we attribute to the effect of a larger dataset with larger point clouds in the original high-dimensional space. Note that the number of edges is directly impacted by the number of nearest neighbors (k) used for the construction of the graph, with a higher k implying a higher number of edges. We attribute this to the increased likelihood of forming well-defined and connected clusters within larger datasets. As the size of the graph increases, there is a greater potential for distinct community structures to emerge. Using the graph-level statistics, we see that graphs that exhibit higher \textbf{assortativity} (coefficient 0.51), \textbf{transitivity} (0.44) and \textbf{local clustering} (0.39) correlate positively with the V-measure. This is also true for \textbf{homophily} (0.66), which encompasses the above properties and the underlying intrinsic structure of the graph. We find that the higher the \textbf{Pagerank}, the lower the V-measure (correlation -0.29). This could be due to the fact that documents with a higher Pagerank score are more general and contain more connections, which leads to scattered cluster structures and removes the homogeneity and completeness of the clusters.

\section{Conclusion and further work}
\label{sec:conclusion}

The main contribution of this work is an incremental $k$-NN algorithm that guarantees connectedness of the neighborhood graph and its analysis. In contrast to related approaches, the proposed algorithm achieves connectedness using only the search for nearest neighbours without requiring additional information to ensure the graph's connectivity. As shown in Section \ref{sec:inc-neig-graphs}, the neighborhood graph created with Algorithm \ref{alg:iter-knn} will always be connected. We show (in Section \ref{sec:show-disconnected-knn}) that $k$-NN neighborhood graphs inferred on realistic datasets can exhibit disconnected components for commonly used values of $k$ -- as well as up to $k=20$ (see Section \ref{sec:results}). Furthermore, when compared to standard $k$-NN, clustering performance is significantly improved for low values of $k$ and of comparable performance for higher values of $k$. Although the incremental nature of the Algorithm \ref{alg:iter-knn} makes it sensitive to the ordering of the nodes, we show that the variance of the resulting graphs has little impact on the clustering performance.
In Section \ref{sec:lat_model}, we show that using bigger embedding models for the initial high-dimensional embedding expectedly yields consistently better results when used with our approach. Unlike approaches described in Section \ref{sec:related-work} that show improved clustering performance with the inclusion of MST, the presented incremental approach does not require any global features in order to guarantee connectedness and does not benefit from the addition of global features like MST.

Proposed incremental $k$-NN neighborhood graph construction is simple and effective, but it may be suboptimal in its initial steps. Nodes that are processed at the start of the algorithm have a higher probability of connecting to a large number of weakly related nodes and disproportionally influence the information encoded with a graph. Future work can analyze alternative strategies to approach the early steps of the algorithm and reduce the risk of suboptimal performance. The neighborhood graph produced by the proposed algorithm can be efficiently extended and allows both the addition of new and the deletion of the last added nodes. Those properties were not leveraged in the clustering evaluation as we only evaluated the final clustering performance. Future work can exploit those properties by focusing on applications where new data is continuously added (streaming data) or invalidated.
We have shown that in order to get clustering performance comparable to spectral clustering on $k$-NN neighborhood graph, the nearest neighbor relation does not need to be exact. Future work can explore using approximate $k$-NN search, e.g., using the algorithm described in \cite{malkov2018efficient}. One of the open problems in graph-based approaches to clustering is the recalculation of all intermediate steps when new data points are introduced \cite{Mondal2024}. With the incremental construction of the neighborhood graph, we make an initial step in addressing this issue, as the existing neighborhood graph can be reused and easily extended with new data.
This property can be readily leveraged in combination with backpropagation-based eigen decomposition \cite{10.5555/3454287.3454571}, or a variant of efficient update of eigenvectors \cite{DHANJAL2014440}.
Future work should apply those approaches and directly quantify the benefits of an incremental approach to data with a temporal dimension. Such evaluation was done for temporal community detection in \cite{Sattar2023}, where authors specifically underline the necessity for new scalable graph algorithms for temporal data -- a challenge for which this work provides a readily applicable solution through its approach to incremental neighborhood graph construction. 
\textbf{Availability} Code and experimental results available upon acceptnace.

\subsubsection*{Acknowledgments}
The work was partially supported by the Slovenian Research and Innovation Agency (ARIS) core research programmes	P2-0103  and P6-0411, as well as the projects EMMA (L2-50070), and LLM4DH (GC-0002). The work of BK was supported by a young researcher grant PR-12394. The work was also supported by the EU through ERA Chair grant no. 101186647 (AI4DH).

\bibliography{bibliograph}

\begin{appendices}

\section{Graph statistics used}
\label{app:graph-statistics}

\begin{itemize}
 \item \textbf{Graph density} measures how close the graph is to being a complete graph. It is defined as:
 $$
 \text{density}(G) = \frac{2|E|}{|V|(|V| - 1)} \\
 $$

 \item \textbf{Assortativity} refers to the tendency of nodes in the graph to be connected to other nodes that are similar in some way. In degree assortativity, it is quantified by the Pearson correlation coefficient of the degrees between pairs of connected nodes:
 \[
 r = \frac{L^{-1} \sum_i j_i k_i - [L^{-1} \sum_i \frac{1}{2}(j_i + k_i)]^2}{L^{-1} \sum_i \frac{1}{2}(j_i^2 + k_i^2) - [L^{-1} \sum_i \frac{1}{2}(j_i + k_i)]^2}
 \]
 where \( j_i \) and \( k_i \) are the degrees of the nodes at the ends of the \(i\)th edge, and \(L\) is the total number of edges.

 \item \textbf{Transitivity} measures the overall degree of clustering in the graph, defined as:
 \[
 C = \frac{3 \times \text{number of triangles in the network}}{\text{number of connected triangles of the nodes}}
 \]

 \item \textbf{Local clustering coefficient} measures the degree to which nodes in a graph tend to cluster together. For a given node \(v\) it is defined as
 \[
 C_v = \frac{2|E_v|}{k_v(k_v - 1)},
 \]
 where \( |E_v| \) is the number of edges between the neighbors of \(v\), and \( k_v \) is the degree of \(v\).

 \item \textbf{Pagerank} is a measure of node importance based on the structure of the incoming links and is defined for a node \(i\) as follows:
 \[
 PR(i) = \frac{1-d}{N} + d \sum_{j \in M(i)} \frac{PR(j)}{L(j)},
 \]
 where \(N\) is the total number of nodes, \(M(i)\) is the set of nodes connected to \(i\), \(L(j)\) is the number of outgoing connections to node \(j\), and \(d\) is the damping factor, which is usually set to 0.85.

 \item \textbf{Homophily} is a tendency of individual nodes to connect and link with similar other nodes. It measures the similarity of connected nodes with respect to a specific attribute, in our case, the true clustering label. For a graph \( G = (V, E) \) where each node (document) \( v \in V \) has a \textit{cluster label}, the homophily \( H \) of the graph can be quantified as the proportion of edges that connect nodes with the same label:
\[
H = \frac{|\{(v_i, v_j) \in E : \text{label}(v_i) = \text{label}(v_j)\}|}{|E|},
\]
where \( \text{label}(v) \) denotes the label of the node \( v \) and \( |\{(v_i, v_j) \in E : \text{label}(v_i) = \text{label}(v_j)\}| \) is the number of edges where both documents have the same label.
\end{itemize}

\end{appendices}

\end{document}